\documentclass[letterpaper, 10 pt, conference]{ieeeconf}  

\IEEEoverridecommandlockouts                              

\usepackage{hyperref}
\usepackage{cite}
\usepackage{amsmath,amssymb,amsfonts}
\usepackage{algorithm}
\usepackage{algorithmic}
\usepackage{graphicx}
\usepackage{textcomp}
\usepackage{xcolor}
\usepackage{bm}
\usepackage{todonotes}

\DeclareMathOperator{\arctantwo}{arctan2}

\title{\LARGE \bf APF-PF: Probabilistic Depth Perception for 3D Reactive Obstacle Avoidance}

\author{Shakeeb Ahmad$^1$, Zachary N. Sunberg$^1$ and J. Sean Humbert$^2$
\thanks{This work was supported through the DARPA Subterranean Challenge,
cooperative agreement number HR0011-18-2-0043}
\thanks{$^1$ Department of Aerospace Engineering Sciences, University of Colorado Boulder, CO, \{shakeeb.ahmad, zachary.sunberg\}@colorado.edu}
\thanks{$^2$ Department of Mechanical Engineering, University of Colorado Boulder, CO, sean.humbert@colorado.edu}
}

\begin{document}

\maketitle
\thispagestyle{empty}
\pagestyle{empty}

\begin{abstract}
This paper proposes a framework for 3D obstacle avoidance in the presence of partial observability of environment obstacles. The method focuses on the utility of the Artificial Potential Function (APF) controller in a practical setting where noisy and incomplete information about the proximity is inevitable. We propose a Particle Filter (PF) approach to estimate potential obstacle locations in an input depth image stream. The probable candidates are then used to generate an action that maneuvers the robot towards the negative gradient of potential at each time instant. Rigorous experimental validation on a quadrotor UAV highlights the robustness and reliability of the method when robot's sensitivity to incorrect perception information can be concerning. The proposed perception and control stack is run onboard the UAV, demonstrating the computational feasibility for real-time applications and agile robots. 
\end{abstract}

\section{Introduction}

There has been an increasing interest in the area of active perception over the last decade or two. A recent DARPA challenge \cite{subt} poses a problem of exploration of unknown, unstructured, large-scale and cluttered underground environments. Like many other sub-problems from the challenge, the path planning problem suffers from uncertain and incomplete sensor information due to noise and limited field of view respectively. Moreover, the computation time and the effectiveness are the essential attributes of a planning method. Mapping-based techniques perform well to generate a path according to a global mission strategy. Commonly used mapping tools like OctoMap \cite{hornung13octomap} and Voxblox \cite{oleynikova2017voxblox} have been popular recently, in generating a map representation of an environment using 3D pointclouds. However, mapping is a slow processes. For large environments, planning over such global information contributes to additional computation time overhead.  Moreover, the coarseness of the map being used for planning over large scale environments may not be suitable to detect intricacies in the environment. A fundamental requirement for a mapping-based planner to be effective, is a precise position feedback of the robot. In practice, such estimates are expensive to obtain, especially locally, in terms of both computation and payload. Therefore, there is a high chance that an agile robot, such as an MAV, does not map small and thin obstacles. Such obstacles, like hanging cables and pipes, are inherent to many complex indoor environments like caves, mines and warehouses. Hence, there is a need to use faster and higher resolution raw sensor measurements to guide a robot. Moreover, a full 3D solution is required to navigate through uncertain and complicated structures. The method proposed in this paper, utilizes raw sensor measurements to generate robot actions without relying on a map representation of an environment. \par 

Researches have shown in past that living organisms \cite{warren2001optic} like insects \cite{frye2001fly, borst2002neural} rely on instantaneous optic flow signals to provide visual cues for reactive obstacle avoidance. One component of an optic flow signal is the sense of how far objects are located in an environment. Depth cameras provide this information in the form of depth images relative to their sensors. A depth image stream can have varying noise levels depending on the type of sensors, the calculation methods involved and abrupt reflections from the objects in the environments. This makes it challenging to obtain useful information to feed into most motion planners that demonstrate well in the perfect observability settings. \par

\begin{figure}
    \centering
    \includegraphics[width=\linewidth]{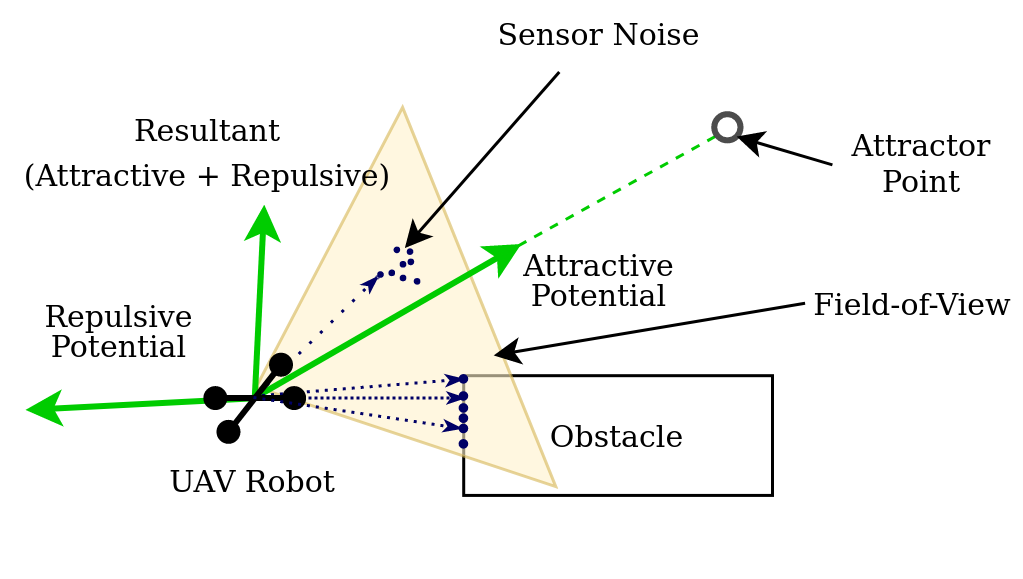}
    \caption{An illustration of the APF-PF problem. The blue dots represent the depth image points as projected in 3D. The blue arrows show the candidates for the obstacle position vector, $\bm{x}_\text{obs}$ (densest dotted line represents the most probable candidate). }
    \label{fig:apf_setup}
\end{figure}

Fragoso \textit{et al.} \cite{fragoso2018dynamically} proposed a way of keeping only a local egocentric map of the environment in the form of a cylinder that updates in real-time through temporal fusion of limited field-of-view visual information. Some other works \cite{matthies2014stereo}, \cite{scherer2009efficient} and \cite{dubey2017droan} propose inflation of depth images using C-Space expansion to facilitate path generation inside a disparity space. Later work by Dubey \textit{et al.} \cite{dubey2018droan} highlights the importance of considering small obstacles like wires, in autonomous robot navigation, by explicitly detecting them using convolutional neural networks. Authors in \cite{matthies2014stereo} and \cite{ahmad2019real} demonstrated closed-loop RRT and LQR based techniques, respectively, for path planning inside a depth image space. Reactive controllers, especially the ones based on Artificial Potential Function (APF) \cite{khatib1986real},  and bio-inspired \cite{conroy2009implementation, humbert2009bioinspired, ohradzansky2020reactive} methods have demonstrated speed and light computational burden. 

\begin{figure}
    \centering
    \includegraphics[width=0.9\linewidth]{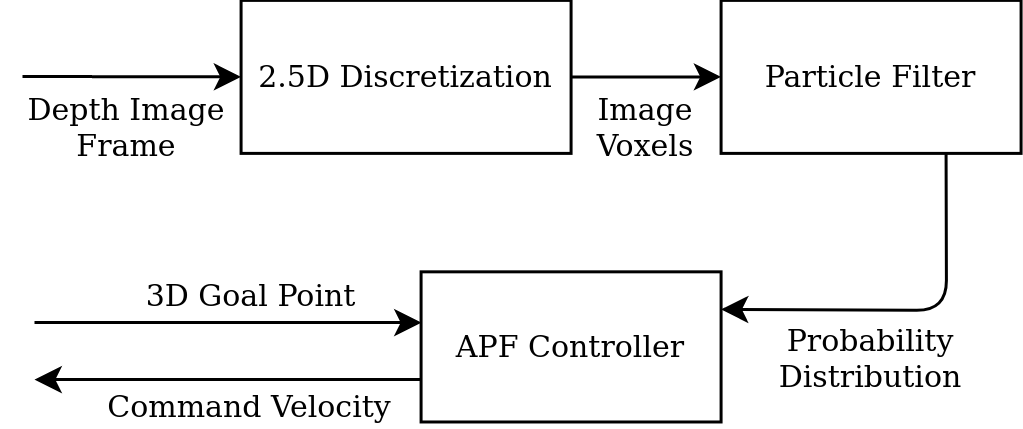}
    \caption{Simplified process flow.}
    \label{fig:basic_flow}
\end{figure}

This paper proposes a solution for the obstacle avoidance problem that
1) generates robot actions directly from an imperfect depth image stream hence providing an end-to-end solution,
2) has a fast update rate to cater for the needs of agile robots,
3) is computationally feasible for robots with limited on-board resources such as MAVs,
4) adapted for robot maneuvers in 3D to make it useful for arbitrarily structured environments.
The perception solution to make such control feasible is the foremost contribution of the paper. A particle filter technique is proposed to probabilistically identify and track regions corresponding to potential obstacles, inside a sequence of depth images. At each time instant, the repulsive potential applied by each candidate for such a region, is weighted according to its probability to compute a total repulsive potential. To close the loop, a user-defined goal point generates an attractive potential and finally the robot is commanded to follow the negative gradient of the resultant potential. Although not uniquely defined in the literature, the term \textit{reactive obstacle avoidance} typically refers to the methods that require only local environment information to generate steering commands away from obstacles and towards the goal, instead of planning paths inside a global map. Our method slightly deviates from the traditional context of reactive obstacle avoidance in that it uses particle filter on a \textit{history} of depth images in order to estimate the local environment information. However, the APF controller in-loop only requires the local information and no map information is available or saved during flight, hence, the method proposed in this paper falls under the umbrella of reactive methods for obstacle avoidance. Fig. \ref{fig:basic_flow} shows the simplified process flow, various components of which are explained in the following sections. \par

The rest of the paper is organized as follows. Section \ref{sec:problem_setup} highlights the problem formulation followed by the description of the filter design, in Section \ref{sec:perception}. Section \ref{sec:apf_control} provides a description of how the control loop is closed with the perception information. Section \ref{sec:results} presents the results and analysis from physical experiments. Finally, Section \ref{sec:conclusion} concludes the discussion.

\section{Problem Setup}
\label{sec:problem_setup}

\textbf{Problem 1:} \emph{Let $\mathcal{X}_\text{free}$ be a set of all collision free robot configurations, $\mathcal{X}_\text{goal} \subset \mathcal{X}_\text{free}$ be the goal region and $\mathcal{A}$ be a set of available robot actions. Given a history of depth images $\bm{I}_{[0 \: t]}$, at any time instant $t$, choose an action $\bm{a}_t \in \mathcal{A}$ such that $\bm{x}_{t+1} \in \mathcal{X}_\text{free}$ and $\bm{x}_{t_f} \in \mathcal{X}_\text{goal}$, at some finite final time $t_f$.}

We aim to solve a well-defined problem of moving a robot from an initial to a final point while avoiding collisions. In the APF approach, the robot behaves as a point charge. The goal point generates an attractive potential for the robot which vanishes as the robot gets closer to the goal. The proximity obstacles are treated as a collection of point charges to generate a repulsive potential to the robot. The robot maneuvers towards the goal, following the negative gradient of the net potential. If $\bm{x}_\text{goal} \in \mathbb{R}^3$ is the vector from the robot's body frame origin to the goal and $\bm{x}_\text{obs} \in \mathbb{R}^3$ is the obstacle's position in the robot's body frame, the attractive and repulsive potential functions, $U_\text{att}: \mathbb{R}^3 \to [0,\infty)$ and $U_\text{rep}: \mathbb{R}^3 \to [0,\infty)$, are given by,

\begin{align}
    \label{eq:att_fun}
    &U_\text{att}(\bm{x}_\text{goal}) =
    \begin{cases}
        \rho_r \xi ||\bm{x}_\text{goal}||,  & ||\bm{x}_\text{goal}|| > \rho_r. \\ 
        \frac{1}{2} \xi ||\bm{x}_\text{goal}||^2,  & ||\bm{x}_\text{goal}|| \leq \rho_r.  
    \end{cases} \\
    \label{eq:rep_fun}
    &U_\text{rep}(\bm{x}_\text{obs}) = 
    \begin{cases}
        \frac{1}{2}\eta(\frac{1}{||\bm{x}_\text{obs}||} - \frac{1}{\rho_0})^2, & ||\bm{x}_\text{obs}|| < \rho_0. \\
        0, & \textnormal{otherwise}.
    \end{cases}
\end{align}
Here, $\rho_0$ defines the sensing horizon over which the repulsive potential is effective. A quadratic potential well is a typical choice for the attractive potential function since it leads to a control law that is linear and all other potentials approximate to quadratic for small perturbations in $\bm{x}_\text{goal}$ \cite{volpe1990manipulator}. However, the conic potential \cite{andrews1983control} provides a constant control law except at the goal \cite{ge2000new}. This behavior is more suitable around the obstacles. The conic attractive function is, therefore, often used close to obstacles and is switched to the parabolic function in close proximity of the goal. In (\ref{eq:att_fun}), $\rho_r$ refers to the radius around the goal where the conic potential function switches to the parabolic potential function. In order to ensure continuity, the conic and quadratic potential gains are set to $\rho_r \xi$ and $\xi$ respectively. \par
An accurate knowledge of an obstacle's location relative to the robot, $\bm{x}_\text{obs}$, is an important variable for the effectiveness of the APF method. Unlike an RGB image, a single channel depth image encodes this information up to an allowable camera resolution. However, depth cameras may have various types of imperfections, making it difficult to rely on the raw output, especially when a system is very sensitive to an incorrect feedback. The following section explains the proposed observer design to estimate such information from a sequence of depth images. Fig. \ref{fig:apf_setup} shows a depiction of the APF-PF problem.

\section{Perception}
\label{sec:perception}

In a recent effort to simulate a No-Depth-Return (NDP) type of noise, authors in \cite{sweeney2019supervised} referred to depth imaging as an area in robotic perception that suffers from a large simulation-reality gap due to scene-dependent noise. This sort of noise is exceptionally difficult to model hence the literature provides little insight into this topic. We base our model on the noise characteristics that adversely affect the detection of potential obstacles in a depth image. These considerations include the density of the clusters, their distances from the camera and their ability to appear, disappear and move during the consecutive image frames. The goal of this problem is to detect candidates for an occupied region inside a depth image. Particle filtering is an appropriate sampling-based approach to the state estimation that can handle large state spaces. \par

The particle filter problem parameters can be defined as a tuple $(\mathcal{S}, \mathcal{O}, \mathcal{T}, \mathcal{Z})$, where $\mathcal{S}$ and $\mathcal{O}$ define the set of all possible states and observations respectively. In our case, $\mathcal{S}$ is a set of \textit{perception} states which refers to the occupied regions inside a depth image. This encodes the information about $\bm{x}_\text{obs}$ without requiring the robot and the obstacles to be mapped in a global frame to generate the repulsive potentials. The probability of a state $\bm{s}_t \in \mathcal{S}$, at any time instant $t$, to transition to a state $\bm{s}_{t+1} \in \mathcal{S}$, at the next time instant, is defined as $\mathcal{T}(\bm{s}_t,\bm{s}_{t+1})$. Similarly, $\mathcal{Z}(\bm{s}_{t+1},\bm{o}_{t+1})$ is the probability of an observation $\bm{o}_{t+1} \in \mathcal{O}$, given a transition to a state $\bm{s}_{t+1} \in \mathcal{S}$. A particle filter outputs a \textit{belief} at every iteration which refers to the probability distribution over all possible states given an observation history. This encodes the relevant information from the entire sequence of observations without needing to explicitly store it.

The following subsections explain various components of the perception problem.
\begin{figure}
    \centering
    \includegraphics[width=\linewidth]{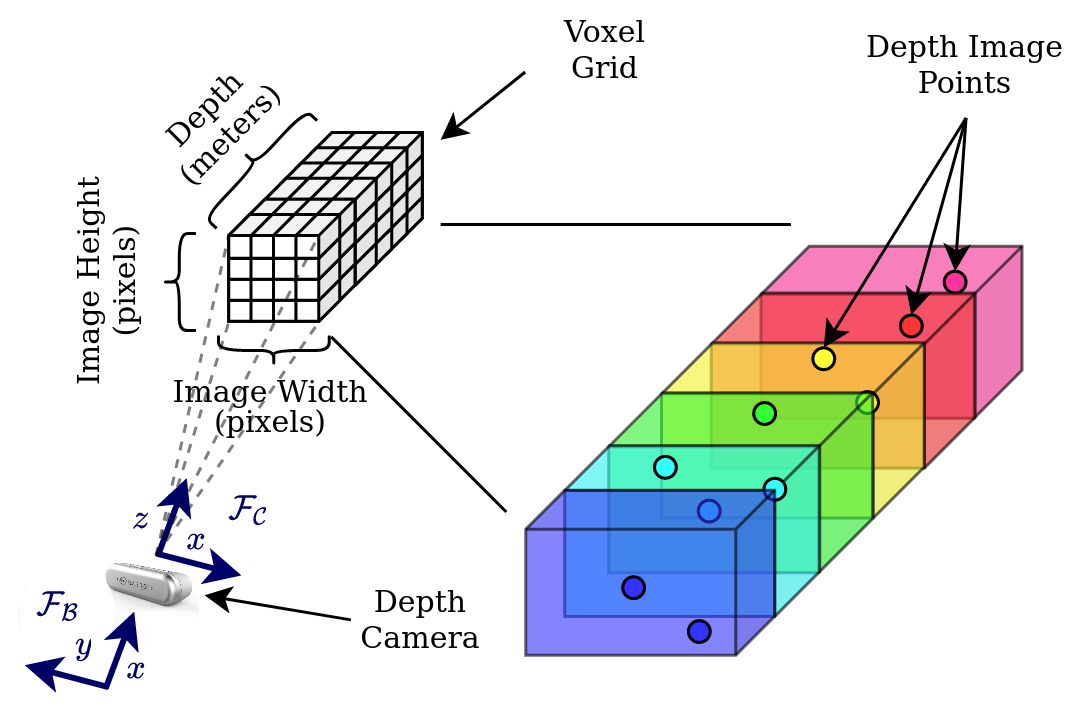}
    \caption{Discretization of a 2.5D depth image. The depth camera's optical frame, $\mathcal{F}_{\mathcal{C}}$, and the robot's body frame, $\mathcal{F}_{\mathcal{B}}$, are shown in the right-handed coordinate system.}
    \label{fig:discretization}
\end{figure}

\subsection{State Space Transition Models}

Any point in an image can correspond to an obstacle in a completely unknown environment and, hence, can potentially be a state. However, at this resolution, the state space size can be large. For a 640$\times$480 image, it is 307200 times the number of possible depths at which an obstacle can be located. This requires the flexibility to discretize such information. We discretize the input image in a 2.5D fashion. This translates to discretization over pixels in the plane parallel to the camera and over depth in the plane perpendicular to the camera. This allows us to make detections directly inside a depth image space in the camera frame, $\mathcal{F}_{\mathcal{C}}$, without projecting the whole image to a 3D space for processing. Moreover, the discretization reduces the state space size up to a user-defined refinement depending on the available computational power. This results in a \textit{voxel} grid where the 2.5D coordinate of each voxel belongs to the state space. When projected to 3D, it forms a pyramid in case of a pinhole camera model, with apex corresponding to the optical center of the camera. This, hence, does not require explicit knowledge of the camera model during the perception process. The 2.5D voxel grid covers the entire sensing horizon (field-of-view). Fig. \ref{fig:discretization} shows such a discretization process, along with the relevant frames of reference. The full state space can be written as,

\begin{align*}
    \mathcal{S} = \{(i, j, k) : i \in \{0,k_w,2k_w, ..., (N_w-1)k_w\}, \\ 
    j \in \{0,k_h,2k_h, ..., (N_h-1)k_h\}, \\
    k \in \{0,k_d,2k_d, ..., (N_d-1)k_d\} \} \cup \{\bm{s}_b\}, 
\end{align*}

where $k_w, k_h \in \mathbb{Z}^+$ are the pixel discretization steps along the width and height of an image and $k_d \in \mathbb{R}^+$ refers to the discretization step along the depth. Similarly, $N_w, N_h$ and $N_d$ are the number of discretization steps along the respective axes. In order for the belief to sum to 1 at each time instant, the state space model also includes a state, $\bm{s}_b$, referring to obstacle-free sensing horizon. The probability of this state, in the belief vector, informs about the likelihood of there being no obstacle in the field-of-view. In the state space model, this state refers to the voxel grid boundary. The idea is that when there is no obstacle inside the sensing horizon, there is always an obstacle surrounding the voxel grid almost surely, and with non-zero probability of transitioning to the voxel grid. \par

The state transition probability is defined as the product of two factors:

\begin{align}
   \label{eq:trans_prob}
   \mathcal{T}(\bm{s}_t,\bm{s}_{t+1}) = P(\bm{s}_{t+1} \mid \bm{s}_t) \: \: \propto \: \: P_\text{move}(\bm{s}_{t+1} \mid \bm{s}_t)P_\text{dist}(\bm{s}_{t+1}) \text{.}
\end{align}
The term, $P_\text{move}(\bm{s}_{t+1} \mid \bm{s}_{t})$, models the relative motion of a scene obstacle by constraining the Manhattan distance $|\text{diag}(k_w, k_h,k_d)^{-1} (\bm{s}_{t+1} - \bm{s}_{t})|$ to follow the normal distribution $2\mathcal{N}(0, \sigma_s)$, with $\sigma_s$ depending on the expected speed of the environment obstacles relative to the robot. This means that an occupied voxel is more likely to transition to a nearby voxel than to a farther one. The states that are close to the voxel grid boundary are highly likely to transition in and out of the field-of-view. The second term, $P_\text{dist}(\bm{s}_{t+1})$, biases the transitions towards the vehicle. Specifically, this bias is defined with a normal distribution centered at zero, $P_\text{dist}(\bm{s}_{t+1}) = P_\text{dist}(s^z_{t+1}) \propto 2\mathcal{N}(0, \sigma_z)$, where $s^z_{t+1}$ is the depth associated with the state $\bm{s}_{t+1}$. The bias helps the perception module to make a more conservative prediction of an obstacle's behavior \textit{i.e.,} coming towards the robot with high probability. Moreover, the closer proximity obstacles have a significantly higher contribution to the repulsive potential than the farther ones. The bias towards approaching obstacles plays an important role in ensuring a safe maneuver for a fast moving robot. 

\subsection{Observation Model}

An observation $\bm{o} \in \mathcal{O}$ is given by,

\begin{align*}
    \bm{o} = \{n_i \in \mathbb{Z}^+ \: : \: i \in \{0, 1, 2, ..., (N_w N_h N_d-1)\}\}.
\end{align*}

In the expression above, $n_i$ represents the number of points contained in the voxel $i$. The observation model refers to the probability of an observation, $\bm{o}_{t+1}$, after transitioning to a state $\bm{s}_{t+1}$ \textit{i.e.,} $\mathcal{Z}(\bm{s}_{t+1}, \bm{o}_{t+1}) = P(\bm{o}_{t+1} \mid \bm{s}_{t+1})$. For any state inside the voxel grid, $\bm{s}_{t+1} \neq \bm{s}_b$, this can be written as,

\begin{align}
    \label{eq:obs_prob}
    &P(\bm{o}_{t+1} \mid \bm{s}_{t+1}) = P(n_0,n_1,...,n_{N_w N_h N_d-1} \mid \bm{s}_{t+1}) = \\
    &\frac{P(\bm{s}_{t+1} \mid  n_0,n_1,...,n_{N_w N_h N_d-1}) P(n_0,n_1,...,n_{N_w N_h N_d-1})}{P(\bm{s}_{t+1})}. \nonumber
\end{align}

Dense cluster of points inside a depth image serves as an important feature to determine that a region is a valid projection of a physical environment object. Therefore, in order to gain confidence on a voxel, the number of points contained within it is observed. Hence, we can conveniently exploit independence of variables.
Equation \ref{eq:obs_prob} can then be reduced to,

\begin{align}
    \label{eq:obs_prob_simplified}
    P(\bm{o}_{t+1} \mid \bm{s}_{t+1}) = \frac{P(\bm{s}_{t+1} \mid n_{\bm{s}_{t+1}}) P(n_0,n_1,...,n_{N_w N_h N_d-1})}{P(\bm{s}_{t+1})},
\end{align}

where $n_{\bm{s}_{t+1}}$ is the number of points contained in the state $\bm{s}_{t+1}$.

Using Bayes's rule on the term $P(\bm{s}_{t+1} \mid n_{\bm{s}_{t+1}})$, and assuming no prior information about $n_i$ for any voxel $i$, 
\begin{align}
    \label{eq:obs_model_1}
    P(\bm{o}_{t+1} \mid \bm{s}_{t+1}) \: \propto \: P(n_{\bm{s}_{t+1}} \mid \bm{s}_{t+1}) = 2\mathcal{N}(k_w k_h,\sigma_o).
\end{align}

Here $k_w k_h$ is the maximum number of points that can be contained in a voxel and depends on the discretization intervals. The tuning parameter, $\sigma_o$, determines the number of points required to gain confidence on any state inside the voxel grid. Low values result in conservative detections where large number of points are required to be certain about a state.

For $\bm{s}_{t+1} = \bm{s}_b$, 
\begin{align}
    \label{eq:obs_model_2}
    &P(\bm{o}_{t+1} \mid \bm{s}_{t+1}) \\
    & = P(n_0 \mid \bm{s}_{t+1})P(n_1 \mid \bm{s}_{t+1})...P(n_{N_w N_h N_d-1} \mid \bm{s}_{t+1}) \nonumber \\
    & \propto \: P(\max_i \: n_i \mid \bm{s}_{t+1}) = 2\mathcal{N}(0, \sigma_n). \nonumber
\end{align}

\begin{figure}
    \centering
    \includegraphics[width = \linewidth]{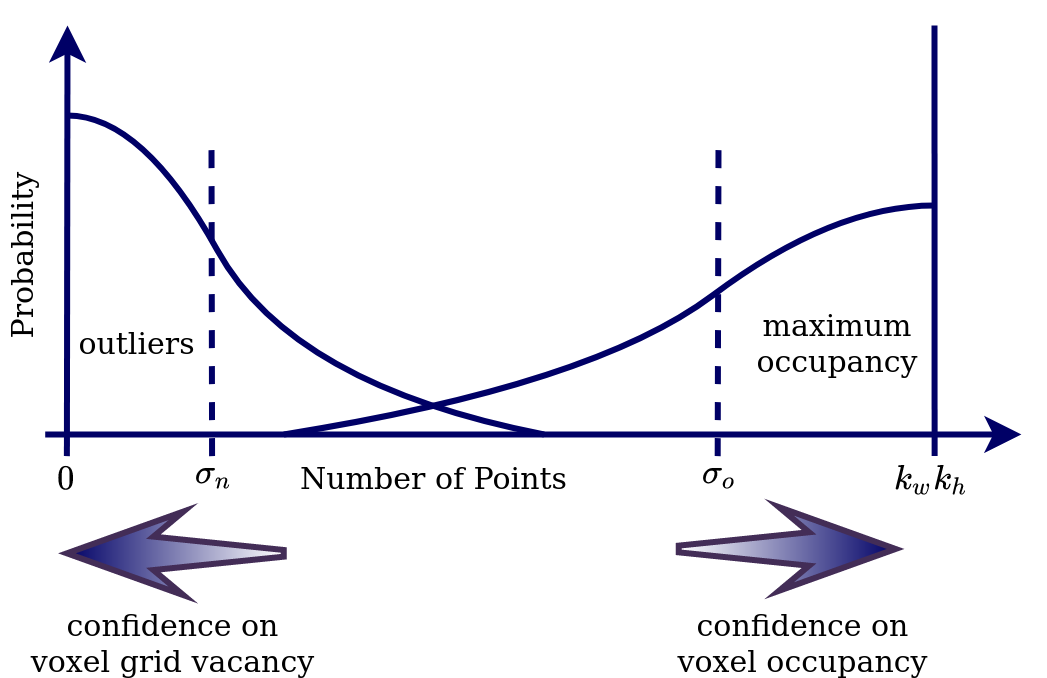}
    \caption{A depiction of the observation model. The horizontal scale represents the number of points contained in a voxel. The number of points towards the left of the scale is more likely to represent outliers. The number of points towards the right of the scale is more likely to represent that the voxel is occupied with an obstacle.}
    \label{fig:obs_model}
\end{figure}

The probability that there is no obstacle inside the voxel grid depends on the number of points in the most populated voxel. This is designed so that, for the system to believe that there is an obstacle in the field-of-view, a significantly large cluster of points needs to exist in atleast one of the voxels. This models the behavior of the outliers in a depth image and helps filter out sparse noise very efficiently. A larger value of $\sigma_n$ helps reject greater number of outliers per voxel. The coexistence of the models defined by (\ref{eq:obs_model_1}) and (\ref{eq:obs_model_2}) is shown in Fig. \ref{fig:obs_model}. By appropriately varying $\sigma_n$ and $\sigma_o$, a variety of different filtering behaviors can be achieved.

\subsection{Particle Filter Algorithm}
The particle filter algorithm is executed on the state and observation models keeping in view the applicability of the approach to systems that require fast and real-time feedback. The transition probabilities, $\mathcal{T}$, depend on the image and discretization parameters which are kept constant throughout the process. Therefore, the first step is to calculate a transition probability lookup table for all possible states. At the start of the run, the belief is initialized with equal number of particles for every state. Every time an image is received, it is first discretized and each voxel is assigned the number of points contained within it. All the particles in the belief vector are then propagated forward in time by randomly drawing, for each particle, a next state from the distribution defined by $\mathcal{T}$. Each propagated particle is then assigned a weight based on the observation model (\ref{eq:obs_model_1}), (\ref{eq:obs_model_2}). Finally, the particles are drawn at random according to the distribution of weights \cite{kochenderfer2015decision}. The probability distribution over the entire state space given the history of depth images, can be computed from the belief vector at any time step. Algorithm \ref{algo:particle_filter} summarizes the estimation process. In the particle filter formulation, the belief about a state is represented by the number of particles associated with that state. For the rest of the paper, the posterior, $P(\bm{s}_t \mid \bm{o}_0, \bm{o}_1, \bm{o}_2, ..., \bm{o}_t \in \mathcal{O})$, is represented by the normalized belief vector $\bm{b}_t$ of length equal to the total number of states. 

\begin{algorithm}[t]
\caption{Particle Filter Estimation}
\label{algo:particle_filter}
    \begin{algorithmic}
        \STATE $b \leftarrow \mathcal{U}(0,N)$
        \STATE $\mathcal{T} \leftarrow $ \textnormal{generate transition probability table given image dimensions and discretization intervals}
        \STATE \textbf{repeat}
        \STATE \quad $b \leftarrow \text{update\_belief}(\text{depth image}, b)$
        
        \STATE \textbf{function} update\_belief(depth image$, b$)
        \STATE \quad $b' \leftarrow \emptyset$
            
        \STATE \quad $\textbf{for}\: i \leftarrow 1 \: \textbf{to} \: N_w N_h N_d-1$
        \STATE \quad \quad $n_i \leftarrow$ number of points inside voxel $i$
            
        \STATE \quad $\textbf{for} \: i \leftarrow 1 \: \textbf{to} \: |b|$
        \STATE \quad \quad $s_i \leftarrow$ random state in $b$
        \STATE \quad \quad draw $k$ with probability proportional to $\mathcal{T}(s_i,.)$
        \STATE \quad \quad $w(i) \leftarrow$ observation\_weight($n_k$)
        \STATE \quad \quad $s'_i \leftarrow s_k$ 
            
        \STATE \quad $\textbf{for} \: i \leftarrow 1 \: \textbf{to} \: |b|$
        \STATE \quad \quad draw $k$ with probability proportional to $w$
        \STATE \quad \quad Add $s'_k$ to $b'$
        \STATE \quad \textbf{return} $b'$
    \end{algorithmic}
\end{algorithm}

\section{APF Control}
\label{sec:apf_control}

The mapping between a pixel inside a depth image to its 3D coordinates in the camera's optical frame, $\mathcal{F}_{\mathcal{C}}$, is defined by the camera's projection model. For a pinhole camera model, it is governed by the following set of equations,

\begin{align}
    \label{eq:pinhole_cam_proj}
    &s^x_t = p^x_t (f_x / p^z_t) + c_x \\
    &s^y_t = p^y_t (f_y / p^z_t) + c_y \nonumber \\
    &s^z_t = p^z_t, \nonumber
\end{align}
where the intrinsics are defined by the camera's focal lengths, $f_x$, $f_y$, and the location of its optical axis, $(c_x, c_y)$, in pixels. The coordinates of a voxel $\bm{s}_t \in \mathcal{S}$ at time instant $t$ are referred to as $(s^x_t, s^y_t, s^z_t)$, with the projected 3D point being $\bm{p}_t = (p^x_t, p^y_t, p^z_t) \in \mathbb{R}^3$. Using (\ref{eq:att_fun}), (\ref{eq:rep_fun}), (\ref{eq:pinhole_cam_proj}), and a belief, $\bm{b}_t$, the net potential, $U_\text{net}: \mathbb{R}^3 \times \mathbb{R}^3 \to [0,\infty)$, on the robot is found as,

\begin{align}
    U_\text{net}(\bm{x}_\text{goal}, \bm{x}_\text{obs}) = \sum_{\bm{s} \in \mathcal{S}} (U_\text{att}(\bm{x}_\text{goal}) + U_\text{rep}(\bm{R}_{\mathcal{C}}^{\mathcal{B}} \bm{p}_t)) b_t(\bm{s}).
\end{align}
where, $\bm{R}_{\mathcal{C}}^{\mathcal{B}}$ is the transformation from the camera frame, $\mathcal{F}_{\mathcal{C}}$, to the robot's body frame, $\mathcal{F}_{\mathcal{B}}$, and $b_t(\bm{s})$ is the belief accociated with a state $\bm{s}$ at time instant $t$ when the potential is computed.

\section{Implementation and Results}
\label{sec:results}

For the ease of integration on an actual robot, we program the perception and control stack in Robot Operating System (ROS) C++ environment. The first subsection examines filter's performance to detect a thin cable in the presence of varying levels of added noise. The second subsection shows results from the flight testing on a real quadrotor UAV robot. For both of the tests, an Intel Realsense D435 \cite{realsense} camera is used to obtain the depth information. The image resolution is set to 640$\times$480 pixels.

\subsection{Thin Cable Test}

\begin{figure}[ht]
    \centering
    \includegraphics[width=\linewidth]{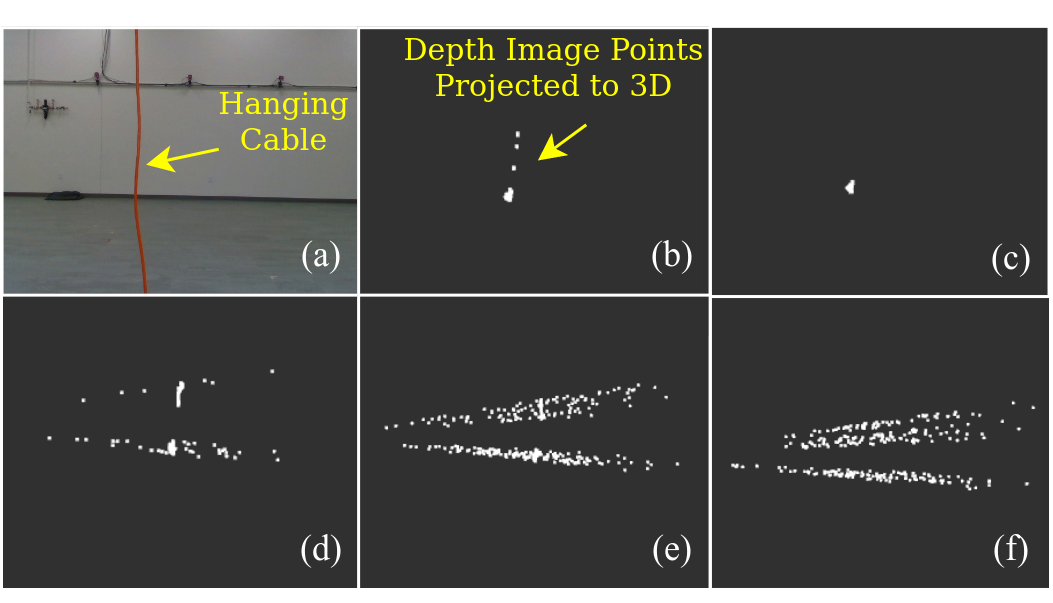}
    \caption{A thin cable hanging infront of the robot. (a) First person view of the cable from the robot. (b),(c) Pointcloud from the side view of the cable, two instances shown. (d) Pointcloud from the side view of the cable with gaussian noise of 0.2 m added to every $10^{th}$ pixel. (e) Pointcloud from the side view of the cable with gaussian noise of 0.2 m added to every other pixel. (f) Pointcloud from the side view of the cable with gaussian noise of 0.2 m added to every pixel. }
    \label{fig:static_cable_rviz}
\end{figure}

Thin and small obstacles pose challenge to flying robots since they are extremely difficult to be mapped. Hanging cables and wires serve as one of the worst case scenarios for this category of obstacles. An extension cord, 8 mm in diameter, is chosen to put filter's performance to test while keeping the robot static. Fig. \ref{fig:static_cable_rviz}(a) shows the RGB image of the hanging cord as seen by the forward facing camera on the stationary robot. Figures \ref{fig:static_cable_rviz}(b),(c) show two snapshots of the depth image when projected to the robot's body frame coordinates. Each 3D point in the figures corresponds to a pixel in the depth image. The maximum number of points an object can occupy in a 640$\times$480 image is 307,200. However, the camera only manages to see less than 300 points on the cord, diminishing to far less than that in many frames.  \par

\begin{figure*}
    \hspace{-24 mm}
    \includegraphics[width=1.24\linewidth]{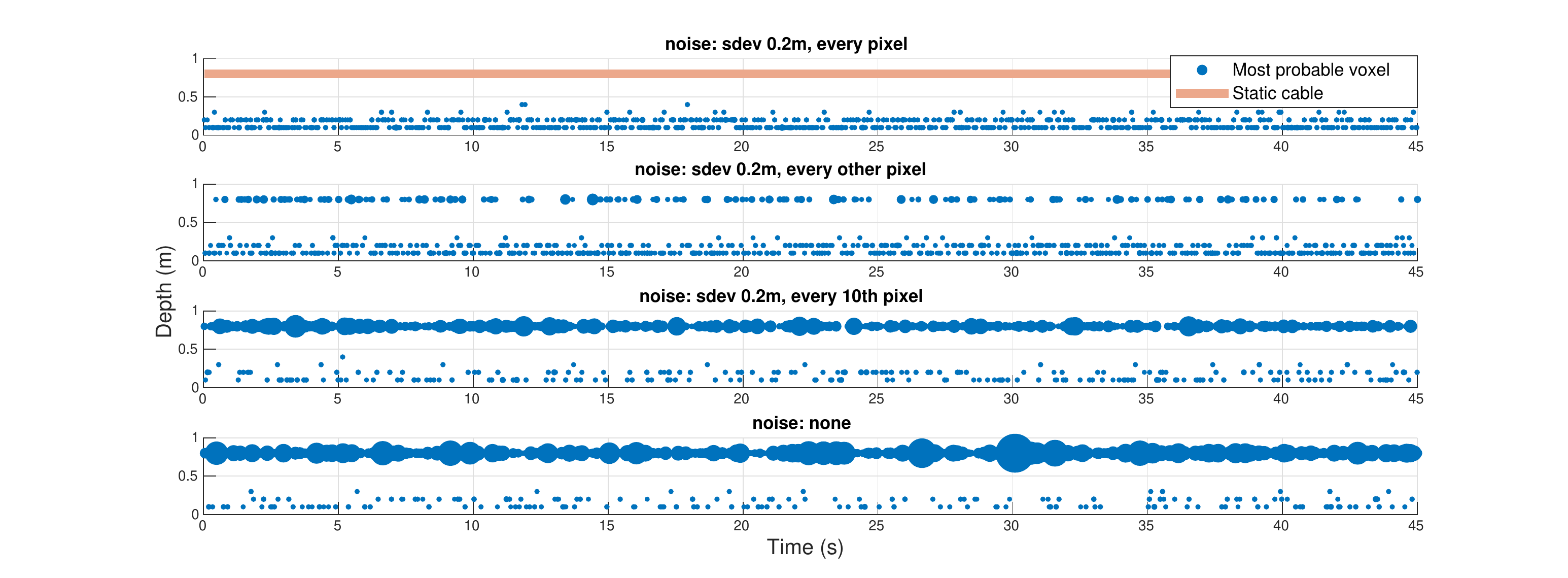}
    \caption{The most probable occupied voxel location over time for a thin static cable test. The thickness of points is proportional to their respective probabilities.}
    \label{fig:static_cable_plot}
\end{figure*}

\begin{figure}
    \centering
    \includegraphics[width=\linewidth]{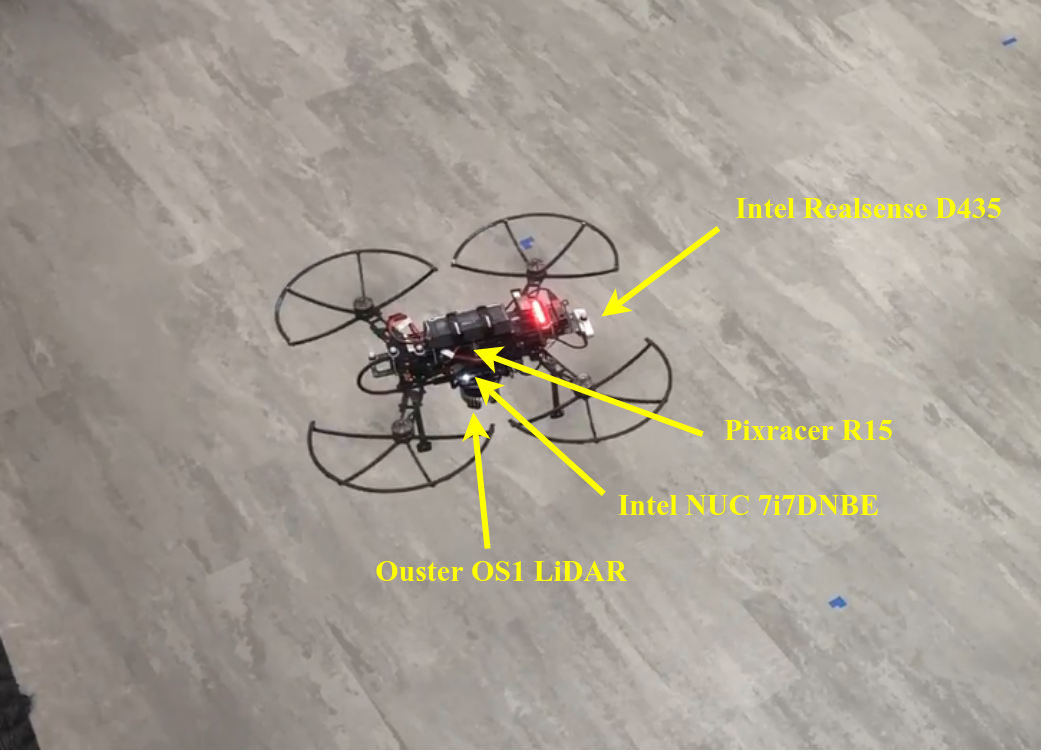}
    \caption{An aerial view of the flying quadrotor UAV.}
    \label{fig:acc_uav}
\end{figure}
In three different tests, we added gaussian noise of standard deviation 0.2 m to the depth associated with every pixel, every other pixel and every 10$^{th}$ pixel in every input image. A fourth test is performed as a control without any noise. The discretization intervals are chosen to be  $k_w = 50, k_h = 50, k_d = 0.1$. The state transition and observation models' covariance parameters are set to $\sigma_s = 8, \sigma_z = 0.4, \sigma_o = 300, \sigma_n = 1$. Additionally, 20,000 particles are used for particle filter estimation. In the APF-PF framework, the repulsive potential applied by each depth image voxel is weighted according to its probability. The probabilities, and hence the influence of the repulsive potentials generated by different voxels, are relative. Therefore, observing the most probable voxel provides a useful insight into the perception performance while keeping the illustration legible. Fig. \ref{fig:static_cable_plot} shows the depth of the most probable voxel transitioning over time for all four noise levels. The size of each vertex in the plot is set proportional to the belief of the corresponding voxel. When the noise is added to every pixel on the cable, there is no information in the image and hence the detections do not necessarily correspond to any physical object in the scene. A bias can be seen towards closer distances due to $P_\text{dist}(\bm{s}_{t+1})$ term in (\ref{eq:trans_prob}).  As the noise gets less dense, the detections appear to happen where the cord is located, with increasing confidence as the density of noise decreases. Like any other observer, there is a performance trade-off if an object of interest resembles the expected sensor noise, as is the case with the cord. Therefore, even with no added noise, the valid detections get less certain during some time instances. This is also because in some frames, for instance Fig. \ref{fig:static_cable_rviz}(b), the structure of the cord disappears to very less number of points before reappearing again in the depth image. The filter does reasonably well in keeping track of the object during those instances. Nevertheless, if the objects of interest in the depth images are more observable or if the noise is more sparse, $\sigma_o$ and $\sigma_n$ can be varied to get more accurate detections. 

\begin{figure}
    \centering
    \includegraphics[width=\linewidth]{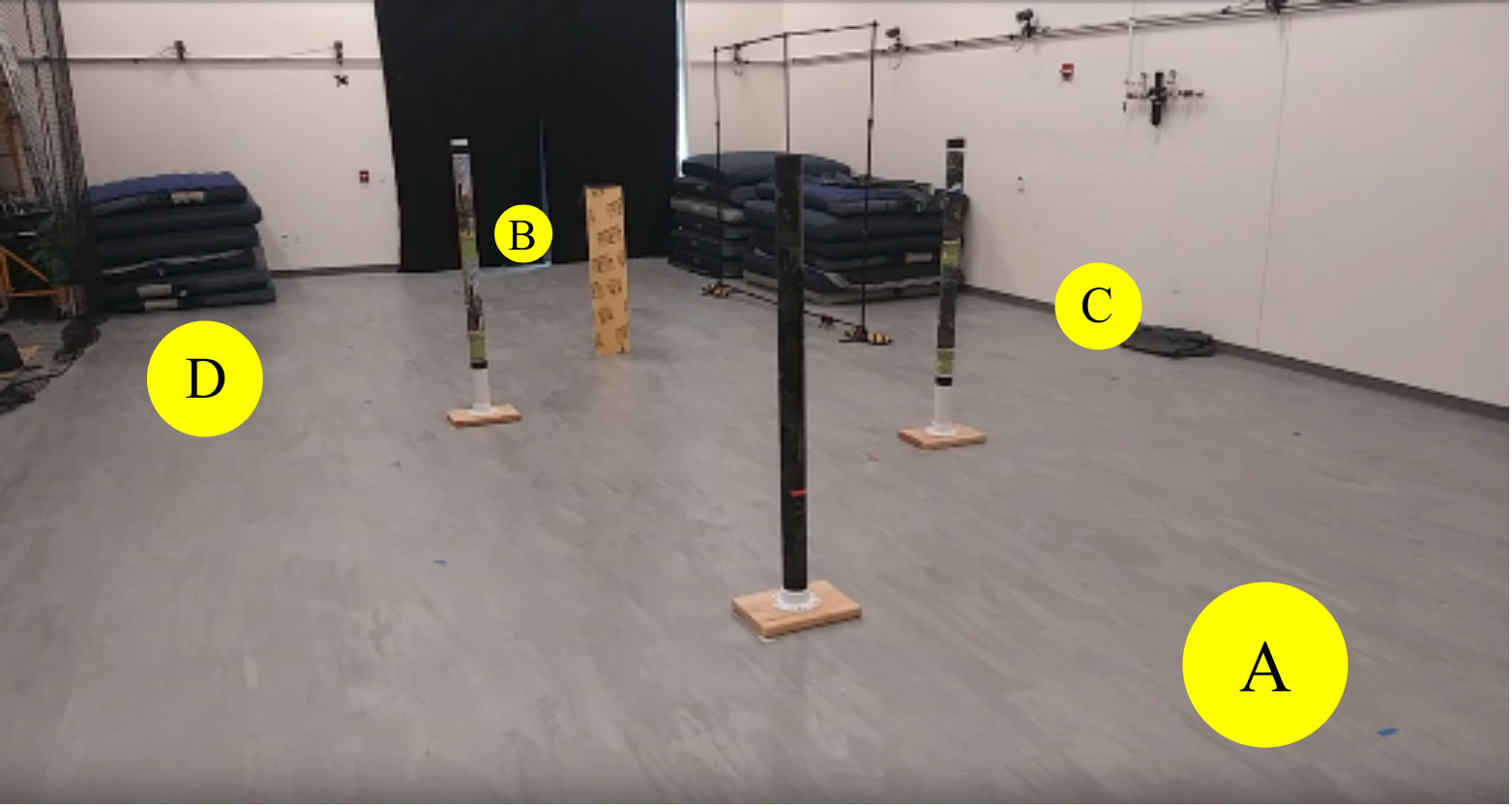}
    \caption{The high bay flight area.}
    \label{fig:flight_area}
\end{figure}
\begin{figure*}
    \hspace{-25.5 mm}
    \includegraphics[width=1.25\linewidth]{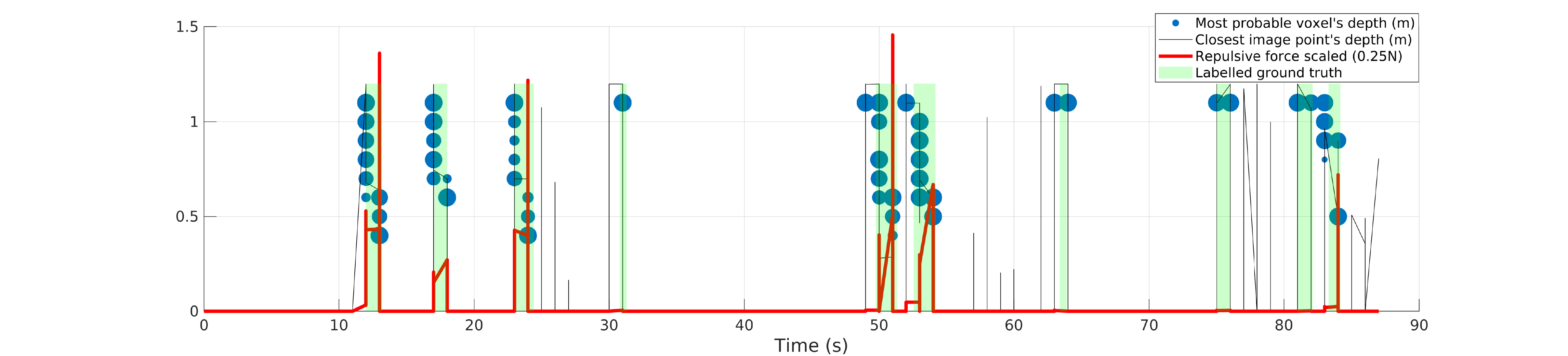}
    \caption{A closeup of one A-B, B-C and C-D run. The green shaded regions represent carefully hand-labelled intervals during which a physical object was inside the camera's field-of-view.}
    \label{fig:flight_rep_plot}
\end{figure*}

\subsection{Physical UAV Experiments}

We performed several flight tests in a high bay area approximately 17 m long and 8 m wide. Results from all of the flight tests are shown in this section. The quadrotor UAV (Fig. \ref{fig:acc_uav}) is built out of a 540 mm airframe equipped with a forward facing Intel Realsense D435 depth camera. The localization estimates on the UAV are provided by a 3D Ouster LiDAR \cite{os1lidar, hess2016real} along with a PX4-based flight controller \cite{pixracer}. The Ouster LiDAR being used has a maximum range of approximately 100 m, a resolution of 1024$\times$64 with horizontal and vertical field-of-views of 360 and 45 degrees, respectively. This makes it very well-suited for localization, but the sparseness of its depth image and the performance degradation at close ranges motivates us to use the Intel Realsense for small obstacle avoidance. Since, we are relying on one camera for these experiments, its utility is increased by adding the non-holonomic constraints on the robot in the $x-y$ plane of its body frame, $\mathcal{F}_\mathcal{B}$. The steering command is generated so that the UAV faces the negative gradient of potential, $ \bm{\nu} = -\nabla U_\text{net} / | \nabla U_\text{net} |$, defined in $\mathcal{F}_\mathcal{B}$.

\begin{align}
    v_t^{\psi} &= \frac{1}{\pi}v^{\psi}_{\max}(\arctantwo (\nu^y,\nu^x)), \\
    v_t^{x} &= v^{x}_{\max}\cos(\arctantwo (\nu^y,\nu^x)), \nonumber \\
    v_t^{z} &= v^{z}_{\max} \nu^z, \nonumber
\end{align}  
where $v_t^x$, $v_t^z$, and $v_t^{\psi}$ are the forward and vertical velocities and the yaw rate in $\mathcal{F}_\mathcal{B}$, respectively. The maximum velocities are regulated by $v^x_{\max}$, $v^z_{\max}$, and $v^{\psi}_{\max}$. Additionally, $\nu^x$, $\nu^y$, and $\nu^z$ are the $x$, $y$ and $z$ components of $\bm{\nu}$. The APF method inherently assumes a holonomic vehicle model. However, by setting $v^{\psi}_{\max}$ sufficiently larger than $v^x_{\max}$, the vehicle's translation closely approximates the holonomic behavior by making yaw to converge much faster than any translation component. In order to avoid the forward velocity to be negative, $v_t^{x}$ is lower bounded by zero. \par

\begin{figure}
    \centering
    \includegraphics[width=\linewidth]{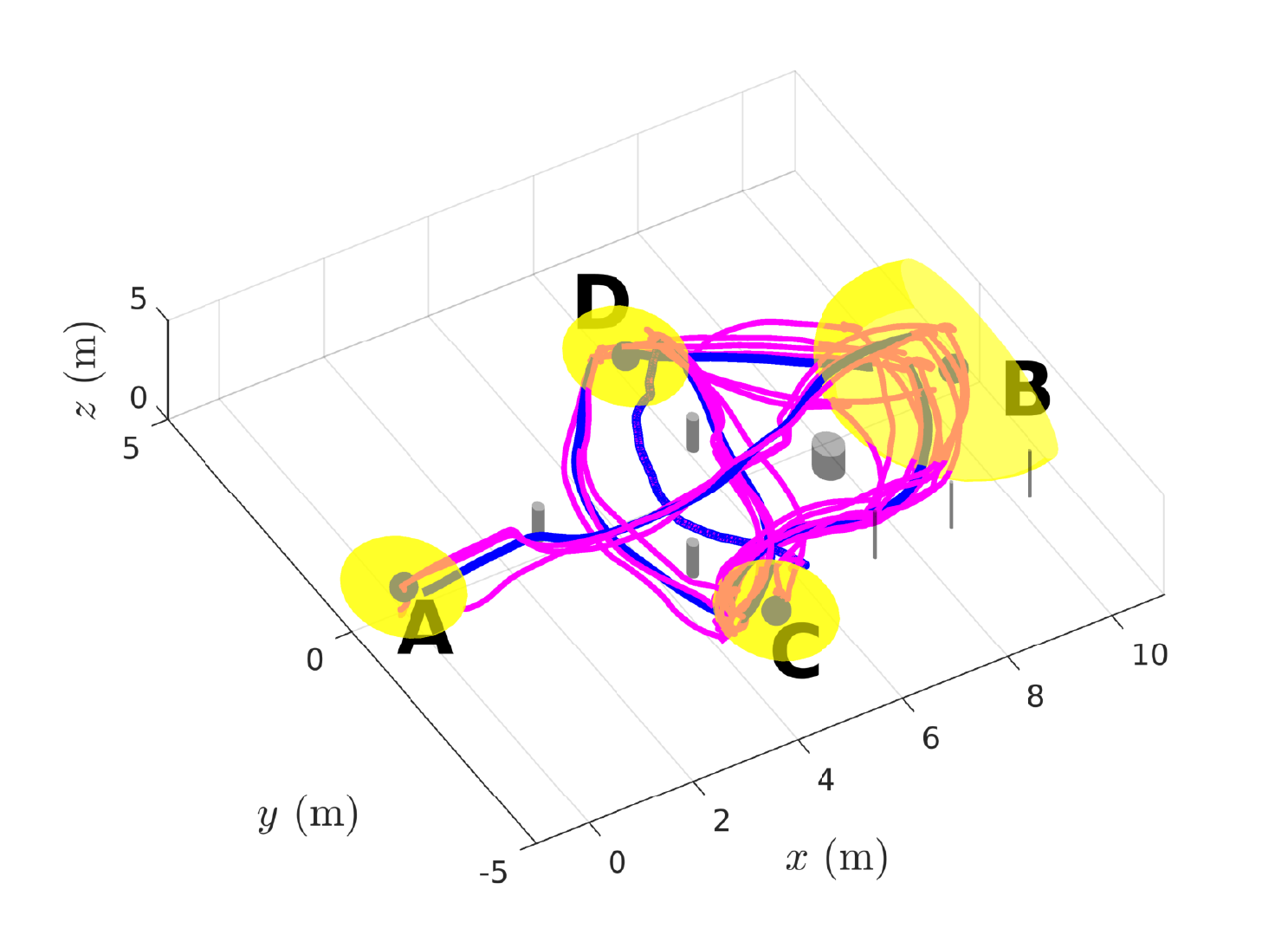}
    \caption{Paths followed by the UAV for multiple A-B, B-C, C-D, D-C and D-B flights. Yellow spheres show the regions A, B, C and D. The blue dots represent the attractor points. The paths followed by the UAV are shown in magenta. For each flight, the UAV takes a different route depending on the initial conditions. The average path over each route is shown by a blue line. }
    \label{fig:flight_path_plot}
\end{figure}

\begin{figure}
    \centering
    \includegraphics[width=0.9\linewidth]{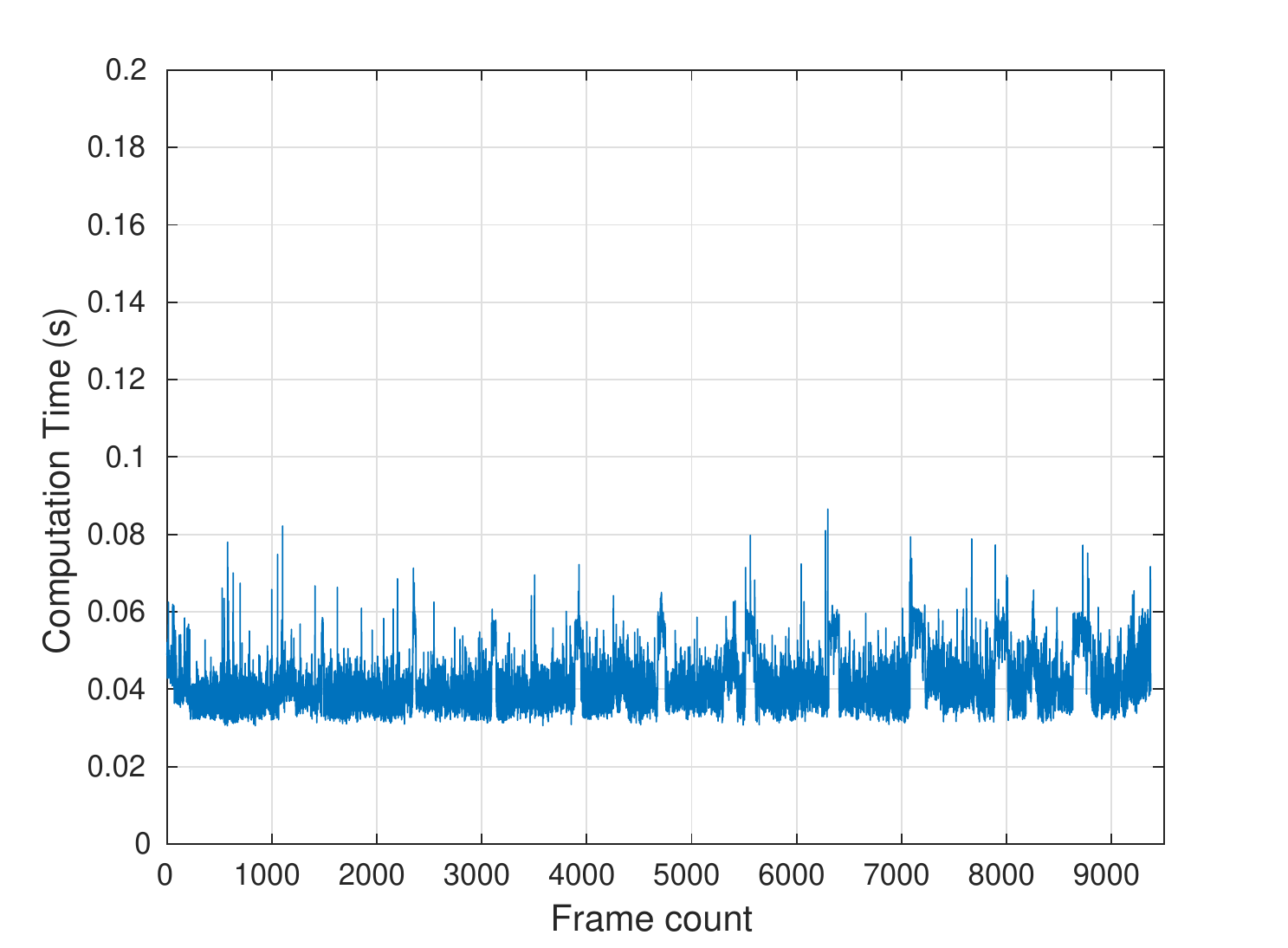}
    \caption{Compute time over several minutes long flight data.}
    \label{fig:flight_compute_plot}
\end{figure}

The covariance parameters, $\sigma_o$ and $\sigma_n$, for the particle filter, are set to 170 and 60 respectively, while keeping all other parameters the same. The control parameters are used as, $\rho_r = 0.5$, $\xi = 0.4$, $\eta = 1.1$, $v^{x}_{\max} = 0.6$, $v^{z}_{\max} = 0.6$, $v^{\psi}_{\max} = 1.0$. Fig. \ref{fig:flight_area} shows the setup with randomly placed obstacles of different reflectivity and of diameters, 270 mm, 90 mm and 22 mm in the course. Each UAV flight path starts from one of the four regions A, B, C or D and ends in one of the three regions B, C or D. The attractor point corresponding to each region is located at its center. The combined duration of the flight tests is approximately 8 min with more than 150 m of distance traveled by the UAV safely. The UAV performs 3 A to B (A-B), 9 B to C (B-C), 8 C to D (C-D), 1 D to C (D-C) and 6 D to B (D-B) flights. The paths followed by the UAV for all the flights, as reported by the LiDAR odometry, with the locations of the obstacles and attractors are shown in Fig. \ref{fig:flight_path_plot}. For the purpose of further analysis, Fig. \ref{fig:flight_rep_plot} shows the performance of the proposed method over one randomly chosen A-B-C-D run. If the APF method was applied directly on the raw sensor measurements, the area of lowest depth would make the most contribution to the repulsive potential. In that case, even low levels of noise can prove highly compromising to system's safety especially if such an area corresponds to noise. The plot shows the depth of the most probable voxel compared against the depth of the closest point inside the raw depth image, at each time instant. The magnitude of the repulsive force is overlaid on the plot. In order to highlight the frequency of raw sensor noise, we hand-labelled the sections in the plot where the UAV had a physical object inside its field-of-view. These sections are shaded green in the figure. At various instances throughout the flight path, the UAV perceived noise that did not belong to any physical object. The repulsive force magnitude is a direct measure of the vehicle's response to those observations. Lastly, the computation time for the complete software stack, over several thousand frames, is shown in Fig. \ref{fig:flight_compute_plot}.

\section{Conclusion}
\label{sec:conclusion}
The paper proposed a perception technique to effectively use an APF-based motion planner on a raw depth image stream. Unlike popular map-based planners, the method is well suited for real-time maneuvers in agile robots especially when precisely building a map is not feasible. The technique is shown to work effectively to perceive and avoid small and thin obstacles from a sequence of noisy depth inputs. The noise level, density and frequency varies with lighting conditions, reflectivity of the surrounding objects, and other environment artifacts. Moreover, depth from stereo is governed by different optics and computation methods than that of a Time-of-Flight (TOF) technology. The method proposed in this paper provides a flexible probabilistic framework to filter out various noise types by carefully tuning the relevant parameters. Finally, the robustness and efficiency of the perception and control stack is demonstrated through physical UAV experiments. \par

A video of all the flight tests is posted at \url{https://youtu.be/oB-WNkrNwR8}.

\section*{Acknowledgment}
This work was supported through the DARPA Subterranean Challenge, cooperative agreement number HR0011-18-2-0043.

\bibliographystyle{./bibliography/IEEEtran}
\bibliography{./bibliography/IEEEexample}

\end{document}